# Learning Non-Lambertian Object Intrinsics across ShapeNet Categories

Jian Shi
Institute of Software, Chinese Academy of Sciences
University of Chinese Academy of Sciences
shij@ios.ac.cn

Yue Dong
Microsoft Research Asia
yuedong@microsoft.com

Hao Su
Stanford University
haosu@cs.stanford.edu

Stella X. Yu
UC Berkeley / ICSI
stellayu@berkeley.edu

## Abstract

*We consider the non-Lambertian object intrinsic problem of recovering diffuse albedo, shading, and specular highlights from a single image of an object.*

*We build a large-scale object intrinsics database based on existing 3D models in the ShapeNet database. Rendered with realistic environment maps, millions of synthetic images of objects and their corresponding albedo, shading, and specular ground-truth images are used to train an encoder-decoder CNN. Once trained, the network can decompose an image into the product of albedo and shading components, along with an additive specular component.*

*Our CNN delivers accurate and sharp results in this classical inverse problem of computer vision, sharp details attributed to skip layer connections at corresponding resolutions from the encoder to the decoder. Benchmarked on our ShapeNet and MIT intrinsics datasets, our model consistently outperforms the state-of-the-art by a large margin.*

*We train and test our CNN on different object categories. Perhaps surprising especially from the CNN classification perspective, our intrinsics CNN generalizes very well across categories. Our analysis shows that feature learning at the encoder stage is more crucial for developing a universal representation across categories.*

*We apply our synthetic data trained model to images and videos downloaded from the internet, and observe robust and realistic intrinsics results. Quality non-Lambertian intrinsics could open up many interesting applications such as image-based albedo and specular editing.*

## 1. Introduction

Specular reflection is common to objects encountered in our daily life. However, existing intrinsic image decomposition algorithms, e.g. SIRFS [3] or Direct Intrinsics (DI)

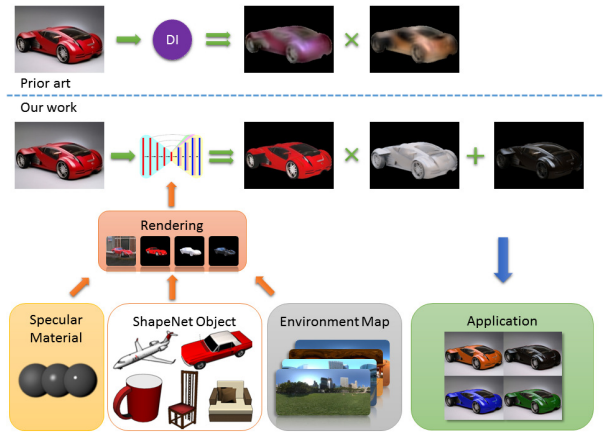

Figure 1: Specularity is everywhere on objects around us and is essential for our material perception. Our task is to decompose an image of a single object into its non-Lambertian intrinsics components that include not only albedo and shading, but also specular highlights. We build a large-scale object non-Lambertian intrinsics database based on *ShapeNet*, and render millions of synthetic images with specular materials and environment maps. We train an encoder-decoder CNN that delivers much sharper and more accurate results than the prior art of direct intrinsics (DI). Our network consistently outperform state-of-the-art especially for non-Lambertian objects and enables realistic applications to image-based albedo and specular editing.

[20], only deal with Lambertian or diffuse reflection. Such mismatching between the reality of images and the model assumption often leads to large errors in the intrinsic image decomposition of real images (Fig. 1).

Our goal is to solve non-Lambertian object intrinsics from a single image. According to optical imaging physics, the old Lambertian model can be extended to a non-

Lambertian model with the specular component as an additive residue term:

$$\text{old}: \text{image } I = \text{albedo } A \times \text{shading } S \quad (1)$$
$$\text{new}: \text{image } I = \text{albedo } A \times \text{shading } S + \text{specular } R \quad (2)$$

We take a data-driven deep learning approach, inspired by DI [20], to learn the associations between the image and its albedo, shading and specular components simultaneously.

The immediate challenge of our non-Lambertian object intrinsics task is the lack of ground-truth data, especially for our non-Lambertian case, and human annotations are just impossible. Existing intrinsics datasets are not only Lambertian in nature, with only albedo and shading components, but also have their own individual caveates. The widely used MIT Intrinsic Images dataset [12] is very small by today's standard, with only 20 object instances under 11 illumination conditions. MPI Sintel [8] intrinsics dataset, used by direct intrinsics, is too artificial, with 18 cartoon-like scenes at 50 frames each. Intrinsics in the Wild (IIW) [5] is the first large-scale intrinsics dataset of real world images, but it provides sparse pairwise human ranking judgements on albedo only, an inadequate measure on full image intrinsic image decomposition.

Another major challenge is how to learn a full multiple image regression task at pixel- and intensity- accurate level. Deep learning has been tremendously successful for image classification and somewhat successful for semantic segmentation and depth regression. The main differences lie in the spatial and tonal resolution demanded at the output: It is full image and 1 bit for classification, more for segmentation and depth regression, most for intrinsics tasks. The state-of-the-art DI CNN model [20] is adapted from a depth regression CNN with a coarse native spatial resolution. Their results are not only blurry, but also with false structures – variations in the output intrinsics out of no structures in the input image. While benchmark scores for many CNN intrinsics models [21, 34, 35, 20, 22] are improving, the visual quality of results remains poor, compared with those from traditional approaches based on hand-crafted features and multitudes of priors [6].

Our work address these challenges and makes the following contributions.

1. New non-Lambertian object intrinsics dataset. We develop a new rendering-based object-centric intrinsics dataset with specular reflection based on ShapeNet, a large-scale 3D shape dataset.

2. New CNN with accurate and sharp results. Our approach not only significantly outperforms the state-of-the-art by every error metric, but also produces much sharper and detailed visual results.

3. Analysis on cross-category generalization. Surprising from deep learning perspective on classification or segmentation, our intrinsics CNN shows remarkable generalization across categories: networks trained only on *chairs* also obtain reasonable performance on other categories such as *cars*. Our analysis on cross-category training and testing results reveal that features learned at the encoder stage is the key for developing a universal representation across categories.

Our model delivers solid non-Lambertian intrinsics results on real images and videos, closing the gap between intrinsic image algorithm development and practical applications.

## 2. Related Work

**Intrinsic Image Decomposition.** Much effort has been devoted to this long standing ill-posed problem [4] of decomposing an image into a reflectance layer and a shading layer. Land and McCann [18] observe that large gradients in images usually correspond to changes in reflectance and small gradients to smooth shading variations. To tackle this ill-posed problem where two outputs are sought out of a single input, many priors that constrain the solution space have been explored, such as reflectance sparsity [28, 30], non-local texture [29], shape and illumination [3], etc. Another line of approaches seek additional information from the input, such as image sequences [33], depth [2, 10] and user strokes [7]. A major challenge in intrinsics research is the lack of ground-truthed dataset. Grosse *et al*. [12] capture the first real image dataset in a lab setting, with limited size and variations. Bell *et al*. [5] used crowdsourcing to obtain human judgements on pairs of pixels.

**Deep Learning.** Narihira *et al*. [21] is the first to use deep learning to learn albedo from IIW's sparse human judgement data. Zhou *et al*. [34] and Zoran *et al*. [35] extend the IIW-CRF model with a CNN learning component. Direct Intrinsics [20] is the first entirely deep learning model that outputs intrinsics predictions, based on the depth regression CNN by [11] and trained on the synthetic MPI Sintel intrinsics dataset. Their results are blurry, with downsampling and convolutions followed by deconvolutions, and poor due to training on artificial scenes. Our work builds upon the success of skip layer connections used in deep CNNs for classification [14] and segmentation [25, 27]. We propose so-called *mirror-links* to forward early encoder features to later decoder layers to generate sharp details.

**Reflectance Estimation.** Multiple images are usually required for an accurate estimation of surface albedo. Aittala *et al*. [1] proposes a learning based method for single image inputs, assuming that the surface only contains stochastic textures and is lit by known lighting directions. Most methods work on homogeneous objects lit by distant light sources, with surface reflectance and environment lighting estimated via blind deconvolution [26] or trained regression networks [25]. Our work aims at general intrinsic image decomposition from a single image, without constraints on material or lighting distributions. Our model predicts spatially varying albedo maps and supports general lighting conditions.

**Learning from Rendered Images.** Images rendered from 3D models are widely used in deep learning, e.g. [31, 19, 13, 23] for training object detectors and viewpoint classifiers. [32] obtains state-of-the-art results for viewpoint estimation by adapting CNNs trained from synthetic images to real ones. ShapeNet [9] provides 330,000 annotated models from over 4,000 categories, with rich texture information from artists. We build our non-Lambertian intrinsics dataset and algorithms based on ShapeNet, rendering and learning from photorealistic images on many varieties of common objects.

## 3. Intrinsic Image with Specular Reflectance

We derived our non-Lambertian intrinsic decomposition equation based on physics-based rendering. Given an input image, the observed outgoing radiance $I$ at each pixel can be formulated as the product integral between incident lighting $L$ and surface reflectance $\rho$ via this rendering equation [16]:

$$I = \int_{\Omega_+} \rho(\omega_i, \omega_o)(N \cdot \omega_i) L(\omega_i) \, d\omega_i. \quad (3)$$

Here, $\omega_o$ is the viewing direction, $\omega_i$ the lighting direction from the upper hemisphere domain $\Omega_+$, and $N$ the surface normal direction of the object.

Surface reflectance $\rho$ is a 4D function usually defined as the bi-directional reflectance distribution function (BRDF). Various BRDF models have been proposed, all sharing a similar structure with a diffuse term $\rho_d$ and a specular term $\rho_s$, and coefficients $\alpha_d, \alpha_s$:

$$\rho = \alpha_d \cdot \rho_d + \alpha_s \cdot \rho_s \quad (4)$$

For the diffuse component, lights scatter multiple times and produce view-independent and low-frequency smooth appearance. By contrast, for the specular component, lights scatter on the surface point only once and produce shinny appearance. The scope of reflection is modeled by diffuse albedo $\alpha_d$ and specular albedo $\alpha_s$.

Combining reflection equation (4) and rendering equation (3), we have the following image formation model:

$$\begin{aligned} I &= \alpha_d \int_{\Omega_+} \rho_d(\omega_i, \omega_o) L(\omega_i) \, d\omega_i \\ &+ \alpha_s \int_{\Omega_+} \rho_s(\omega_i, \omega_o) L(\omega_i) \, d\omega_i = \alpha_d s_d + \alpha_s s_s, \end{aligned} \quad (5)$$

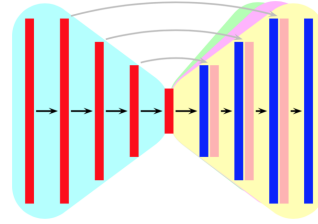

Figure 2: Our mirror-link CNN architecture has one shared encoder and three decoders for albedo, shading, specular components separately. Mirror links connect the encoder and decoder layers of the same spatial resolution, providing visual details. The height of layers in this figure indicates the spatial resolution.

where $s_d$ and $s_s$ are the diffuse and specular shading, respectively. Traditional intrinsics models consider diffuse shading only, by decomposing the input image $I$ as a product of diffuse albedo $A$ and shading $S$. However, it is only proper to model diffuse and specular components separately, since their albedos have different values and spatial distributions. The usual decomposition of $I = A \times S$ is only a crude approximation.

Specular reflectance $\alpha_s s_s$ has characteristics very different from diffuse reflectance $\alpha_d s_d$: Both specular albedo and specular shading have high-frequency spatial distributions and color variations, making decomposition more ambiguous. We thus choose to model specular reflectance as a single residual term $R$, resulting in the non-Lambertian extension: $I = A \times S + R$, where input image $I$ is decomposed into diffuse albedo $A$, diffuse shading $S$, and specular reflectance $R$ respectively.

Our image formation model is developed based on physics based rendering and physical properties of diffuse and specular reflection, and it does not assume any specific BRDF model. Simple BRDF models (*e.g.* Phong) can be used for rendering efficiency, and complex models (*e.g.* Cook-Torrance) for higher photo-realism.

## 4. Learning Intrinsics

We develop our CNN model and training procedure for non-Lambertian intrinsics.

**Mirror-Link CNN.** Fig. 2 illustrates our encoder-decoder CNN architecture. The encoder progressively extracts and down-samples features, while the decoder up-samples and combines them to construct the output intrinsic components. The sizes of feature maps (including input/output) are exactly **mirrored** in our network. We **link** early encoder features to the corresponding decoder layers at the same spatial resolution, in order to obtain local sharp details preserved in early encoder layers. We share the same encoder and use separate decoders for $A, S, R$.

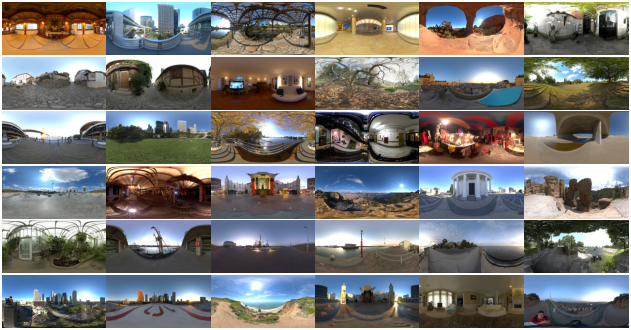

Figure 3: Environment maps are employed in our rendering for realistic appearance, both outdoor and indoor scenes are included. The environment map not only represents the dominate light sources in the scene (*e.g.* sun, lamp and window) but also includes correct information on the surroundings (*e.g.* sky, wall and building). Although a dominate light might be sufficient for shading a Lambertian surface, detailed surroundings provide the details in the specular.

Our mirror links are similar to skip connections in Deep Reflectane Map (DRM) [25] and UNet [27]. However, our goal is entirely different: DRM solves an interpolation problem from high resolution sparse inputs to low resolution dense map outputs in the geometry space, ignoring the spatial inhomogeneity of reflectance, whereas UNet deals with image segmentation rather than image-wise regression.

**Edge sensitive loss.** Human vision is sensitive to edges, however standard loss functions such as MSE treat pixel errors equally. To get more precise and sharp edges, we re-weight pixel errors with image gradients.

**Scale invariant Loss.** There is an inherent scale ambiguity between albedo and shading, as only their product matters in the intrinsic image decomposition. [20] employs a weighted combination of MSE loss and scale-invariant MSE loss for training their intrinsic networks. Scaling ambiguity also exists in our formulation, and we combine these loss functions with our edge-sensitive weighting for training our network.

**ShapeNet-Intrinsics Dataset.** We obtain the geometry and albedo texture of 3D shapes from ShapeNet, a large-scale richly-annotated, 3D shape repository [9]. We pick 31,072 models from several common categories: car, chair, bus, sofa, airplane, bench, container, vessel, *etc*. These objects often have specular reflections.

**Environment maps.** To generate photo-realistic images, we collect 98 HDR environment maps from online public resources[1]. Indoor and outdoor scenes with various illumination conditions are included, as shown in Fig. 3.

**Rendering.** We use an open-source renderer Mitsuba [15] to render objects models with various environment maps and random viewspoints sampled from the upper hemisphere. A modified Phong reflectance model [24, 17] is assigned to objects to generate photo-realistic shading and specular effects. Since original models in ShapeNet are only provided with reliable diffuse albedo, we use random distribution for specular with $k_s \in (0, 0.3)$ and $N_s \in (0, 300)$, which covers the range from pure diffuse to high specular appearance (Fig. 1). We render albedo, shading and specular layers, and then synthesize images according to Equation 5.

**Training.** We split our dataset at the object level in order to avoid images of the same object appearing in both training and testing sets. We use $80/20$ split, resulting in $24,932$ models for training and $6,240$ for testing. All the 98 environment maps are used to rendering $2,443,336$ images for the training set. For the testing set, we randomly pick 1 image per testing model.

More implementation details can be found in the supplementary materials.

## 5. Evaluation

Our method is evaluated and compared with SIRFS [3], IIW [5], and Direct Intrinsics (DI) [20]. We also train DI using our ShapeNet intrinsics dataset and denote the model as DI*. We adopt the usual metrics, MSE, LMSE and DSSIM, for quantitative evaluation.

### 5.1. Synthetic Data

Table 1 shows the numeric evaluation on the synthetic testing set. Our algorithm performs consistently better than others on the synthetic dataset numerically, compared to off-the-shelf solutions, our method provides 40-50% performance gain on the DSSIM error. Also note that, DI*, i.e. DI trained with our dataset, produces second best results across almost all the error metrics, demonstrating the advantage of our ShapeNet intrinsics dataset.

Numerical error metrics may not be fully indicative of visual qualities, *e.g.* the naive baseline also produces low errors for some cases. Figure 4 provides visual comparisons against ground truths.

For objects with strong specular reflectance, e.g. cars,

| ShapeNet intrinsics | MSE | | LMSE | | DSSIM | |
|---|---|---|---|---|---|---|
| | albedo | shading | albedo | shading | albedo | shading |
| Baseline | 0.0232 | 0.0153 | 0.0789 | 0.0231 | 0.2273 | 0.2341 |
| SIRFS | 0.0211 | 0.0227 | 0.0693 | 0.0324 | 0.2038 | 0.1356 |
| IIW | 0.0147 | 0.0149 | 0.0481 | 0.0228 | 0.1649 | 0.1367 |
| DI | 0.0252 | 0.0245 | 0.0711 | 0.0275 | 0.1984 | 0.1454 |
| DI* | 0.0115 | 0.0066 | 0.0470 | 0.0115 | 0.1655 | 0.0996 |
| Ours | **0.0083** | **0.0055** | **0.0353** | **0.0097** | **0.0939** | **0.0622** |
| specular | 0.0042 | | 0.0578 | | 0.0831 | |

Table 1: Evaluation on our synthetic dataset. For the baseline, we set its albedo to be the input image and its shading to be 1.0. The last row lists our specular error.

---

[1] http://www.hdrlabs.com/sibl/archive.html

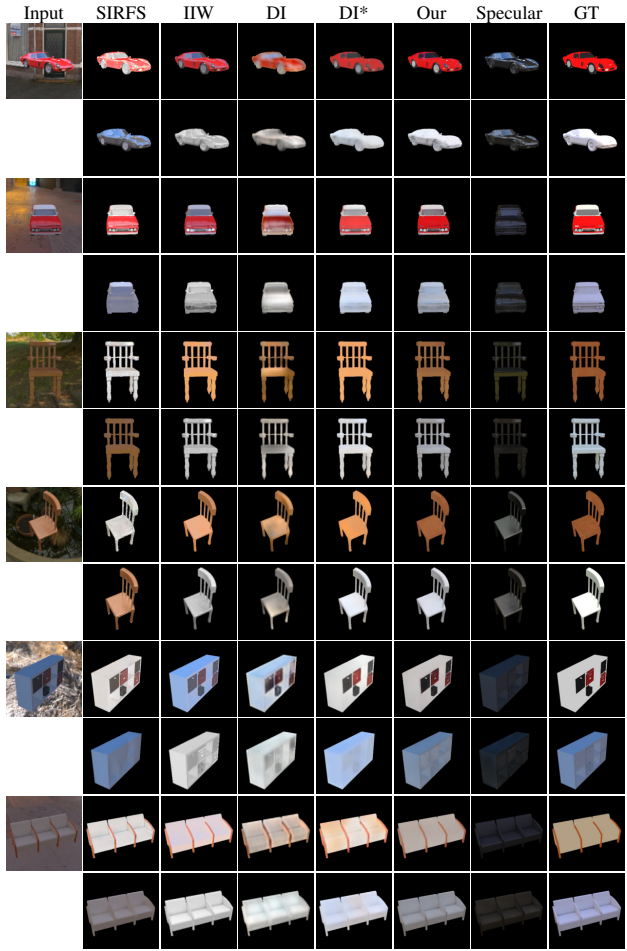

Figure 4: Results for the synthetic dataset. Our baselines include SIRFS, IIW, Direct-Intrinsics with released model by the author (DI), and model trained by ourselves on our synthetic dataset (DI*). The top row of each group is albedo, and the bottom is shading. The *Specular* column shows the ground-truth (top) and our result (bottom). We observe that specularity has basically been removed from albedo/shading, especially for cars. Even for the sofa (last row) with little specular, our method still produces good visual result. See more results in our supplementary material.

| MIT | MSE | | LMSE | | DSSIM | |
|---|---|---|---|---|---|---|
| intrinsic | albedo | shading | albedo | shading | albedo | shading |
| SIRFS | **0.0147** | **0.0083** | **0.0416** | **0.0168** | **0.1238** | **0.0985** |
| DI | 0.0277 | 0.0154 | 0.0585 | 0.0295 | 0.1526 | 0.1328 |
| Ours | 0.0468 | 0.0194 | 0.0752 | 0.0318 | 0.1825 | 0.1667 |
| Ours* | 0.0278 | 0.0126 | 0.0503 | 0.0240 | 0.1465 | 0.1200 |

Table 2: Evaluation on MIT intrinsics dataset.

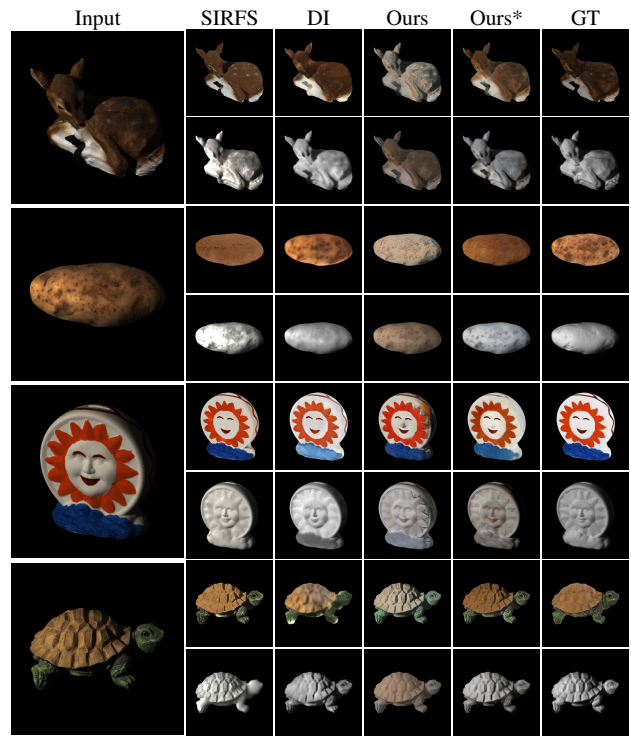

Figure 5: Results on the MIT dataset. *Ours** is our ShapeNet trained model fine-tuned on MIT, with data generated by the GenMIT approach used in DI [20].

specular reflection violates the Lambertian condition assumed by traditional intrinsics algorithms. These algorithms, SIRFS or IIW, simply cannot handle such specular components. Learning-based approaches, DI, DI*, or our method, could still learn from the data and perform better in these cases. For DI, the network trained on our ShapeNet category also has significantly better visual quality, compared with their released model trained on the Sintel dataset. However, their results are blurry, as a consequence from their deep convolution and deconvolution structures without our skip layer connections. Our model produces sharper images preserving many visual details, such as boundaries in the albedo and specular images. Large specular areas on the body of cars are also extracted well in the specular output component, revealing the environment illumination. Such specular areas would confuse earlier algorithms and bring serious artifacts to albedo/shading.

### 5.2. MIT Intrinsics Dataset

We also test the performance of our network on the MIT intrinsics dataset [12]. Unlike our color environment light model designed for common real-world images, the MIT-intrinsics dataset uses a lab capture oriented lighting model with single grayscale directional light source, a scenario that is not included in our synthetic dataset. The light model differences lead to dramatic visual differences and cause domain shift problems on learning based approaches [20]. We also follow [20] to fine tune our network on the MIT dataset.

Table 2 lists benchmark errors and Fig 5 provides sample results for visual comparisons. SIRFS produces the best numerical results, since the pure Lambertian surface reflection and grayscale lighting setup best fits the assumption of such prior-based intrinsics algorithms. Direct intrinsics [20] requires fine tuning to reach similar performance. Our model fine tuned on the MIT dataset produces comparable results as SIRFS and better than DI finetuned on MIT; in addition, our results preserve more details compared to DI.

### 5.3. Real Images

We also evaluate our algorithm on real images as shown in Figure 6. Although our model is trained on a purely synthetic dataset, it produces better results on real images compared to other algorithms, thanks to the realistic rendering that simulates the physical effects of diffuse and specular reflection well and to the generalization properties of our task. Our network produces good quality results on objects never included in our dataset, such as mouse, toy and fruits. In many results such as the car, mouse and potato, specular highlights are correctly estimated, and the corresponding albedo maps are recovered with correct colors. Note that highlight pixels could be so bright that no diffuse colors are left in the input pixels, essentially invalidate many chroma based solutions. Finally, we apply our model to videos frame by frame, and obtain coherent results without any constraints on temporal consistency. Please see our videos and material editing results in supplementary materials.

## 6. Cross-category generalization

Our ShapeNet intrinsics dataset provides semantic category information for each object, allowing in-depth analysis for cross-category performance analysis of our learning based intrinsic image decomposition task. We conduct category-specific training of our network on 4 individual categories which have more than 3,000 objects each: car, chair, airplane and sofa. We evaluate the network on the entire dataset as well as these 4 categories. All these networks are trained with the same number of iterations regardless the number of training data.

Table 3 shows the test performance on individual categories. For almost all the categories, training on the specific dataset produces the best decomposition results on that category. This result is not surprising, as the network always performs the best at what it is trained for. Training with all the datasets leads to a small prediction error increase, at less than 0.02 in the DSSIM error.

Surprisingly, on an input image of an object category (*e.g.* car) that has never been seen during training (*e.g.* chairs), our network still produces reasonable results, with the DSSIM error on-par or better than existing works that designed for general intrinsic tasks (Table 1). Figure 7 shows sample cross-category training and testing results: All models produce reasonable results, confirming cross-category generalization of those models.

**Analysis on generalization.** Our image-to-image regression network always produces the same physical components: albedo, shading and specular maps, unlike classification networks with semantic labels. Although objects in different categories have dramatically different shapes, textures, and appearances, those components have the same physical definitions and share similar structures. Many those commonalities are widely used in previous intrinsics algorithms,*e.g.* shading is usually smooth and grayscale; albedo contains more color variations and specular is sparse and of higher contrast.

Some properties maintain across a certain categories. For example, the Chair and Sofa categories share similar texture (textile and wood), similar albedo features, and similar shapes, thus their cross-category predictions on all three output channels transfer well.

We also observe non-symmetry in Table 3: *e.g.* the network trained on Car produces good results on Airplane, while the network trained with Airplane has relative larger error on Car. This is related to the within-category variations: the car category has more variations in both shapes and textures, providing richer variations that lead to better generalization. This result can also be observed in the benchmarks on the ALL dataset, where the Car-category network produces the best results except the general ALL-

| Albedo | | | | | |
|---|---|---|---|---|---|
| | ALL | Car | Chair | Airplane | Sofa |
| ALL | **0.0939** | 0.1014 | 0.0988 | 0.0893 | 0.0716 |
| Car | 0.1134 | **0.0808** | 0.1379 | 0.1057 | 0.1002 |
| Chair | 0.1181 | 0.1578 | **0.0911** | 0.1166 | 0.0835 |
| Airplane | 0.1201 | 0.1410 | 0.1338 | **0.0757** | 0.0954 |
| Sofa | 0.1131 | 0.1348 | 0.1101 | 0.1067 | **0.0663** |
| Shading | | | | | |
| | ALL | Car | Chair | Airplane | Sofa |
| ALL | **0.0622** | 0.0685 | **0.0549** | 0.0596 | 0.0491 |
| Car | 0.0687 | **0.0579** | 0.0692 | 0.0683 | 0.0592 |
| Chair | 0.0772 | 0.1008 | 0.0561 | 0.0740 | 0.0548 |
| Airplane | 0.0776 | 0.0936 | 0.0738 | **0.0481** | 0.0629 |
| Sofa | 0.0721 | 0.0877 | 0.0594 | 0.0697 | **0.0460** |
| Specular | | | | | |
| | ALL | Car | Chair | Airplane | Sofa |
| ALL | **0.0831** | 0.0866 | 0.0714 | 0.1021 | 0.0730 |
| Car | 0.0953 | **0.0745** | 0.0962 | 0.1214 | 0.0854 |
| Chair | 0.0982 | 0.1162 | **0.0719** | 0.1205 | 0.0800 |
| Airplane | 0.1019 | 0.1115 | 0.0980 | **0.0871** | 0.0939 |
| Sofa | 0.0984 | 0.1115 | 0.0800 | 0.1238 | **0.0673** |

Table 3: Cross-category DSSIM evaluation. Each row corresponds to a model trained on the specific category, and each column corresponds to the result evaluated on the specific category. Reasonably almost all of the lowest errors show up on the diagonal except the shading for chairs. Category-specific training gives better results on the specific category, while the results in the cross-category setting are still comparable and promising.

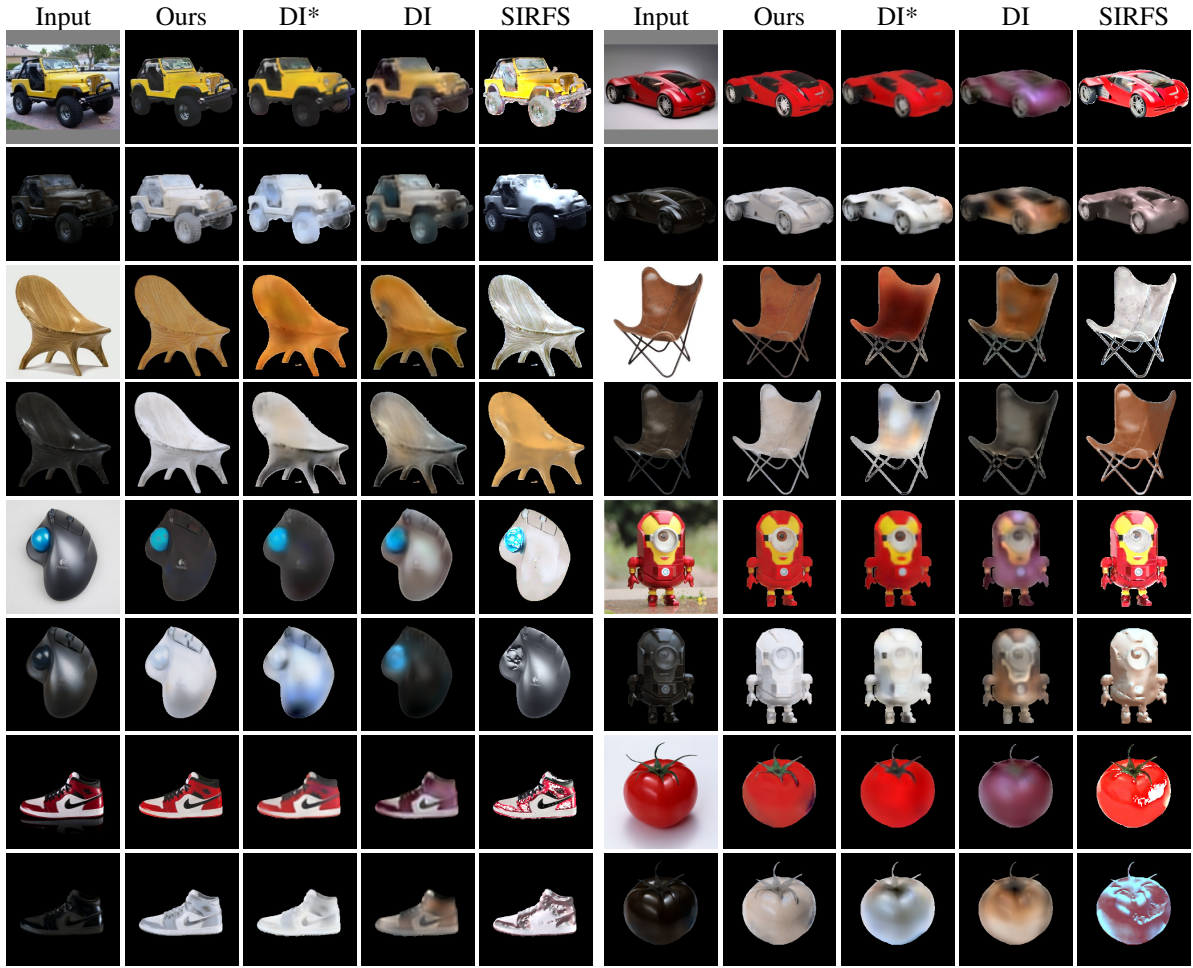

Figure 6: Evaluation on real world images. The first column contains input image (top) and our specular prediction (bottom). For the group of results of an image, the top row gives the predicted albedo and the bottom row gives the shading. We observe that: 1) DI* trained on our dataset produces better results than the publicly released model DI; however, they are still blurry and lose fine details. 2) SIRFS produces erroneous albedo prediction for cases when specular is strong, as specular reflectance is not assumed by SIRFS.

|  | Albedo | | | | Shading | | | | Specular | | | |
| --- | --- | --- | --- | --- | --- | --- | --- | --- | --- | --- | --- | --- |
|  | Car | Chair | Airplane | Sofa | Car | Chair | Airplane | Sofa | Car | Chair | Airplane | Sofa |
| Car | **0.0808** | 0.1379 | 0.1057 | 0.1002 | **0.0579** | 0.0692 | 0.0683 | 0.0592 | **0.0745** | 0.0962 | 0.1214 | 0.0854 |
| Car-Chair | 0.1157 | 0.1303 | 0.1182 | 0.0954 | 0.0769 | 0.0678 | 0.0743 | 0.0598 | 0.0833 | 0.0907 | 0.1215 | 0.0882 |
| Chair-Car | 0.1311 | 0.1111 | 0.1125 | 0.0929 | 0.0873 | 0.0582 | 0.0711 | 0.0573 | 0.1089 | 0.0736 | 0.1235 | 0.0810 |
| Chair | 0.1578 | **0.0911** | 0.1166 | 0.0835 | 0.1008 | **0.0561** | 0.0740 | 0.0548 | 0.1162 | **0.0719** | 0.1205 | 0.0800 |
| Airplane | 0.1410 | 0.1338 | **0.0757** | 0.0954 | 0.0936 | 0.0738 | **0.0481** | 0.0629 | 0.1115 | 0.0980 | **0.0871** | 0.0939 |
| Airplane-Sofa | 0.1502 | 0.1324 | 0.0855 | 0.0938 | 0.0940 | 0.0719 | 0.0546 | 0.0609 | 0.1104 | 0.0932 | 0.0916 | 0.0894 |
| Sofa-Airplane | 0.1349 | 0.1149 | 0.1032 | 0.0723 | 0.0954 | 0.0628 | 0.0703 | 0.0510 | 0.1129 | 0.0829 | 0.1151 | 0.0763 |
| Sofa | 0.1348 | 0.1101 | 0.1067 | **0.0663** | 0.0877 | 0.0594 | 0.0697 | **0.0460** | 0.1115 | 0.0800 | 0.1238 | **0.0673** |

Table 4: Cross-category decoder fine-tuning results. We freeze the encoder component and fine-tune the decoder components on the cross-category setting, to diagnose the important of encoder. Car-Chair stands for the model first trained on cars and then fine-tuned on chairs. Results show that fine-tuning decoder would not bring much performance improvements, if the encoder is trained biasedly. We also observe that cross-category fine-tuning makes little difference when evaluation on a third category, e.g. Car-Chair on Sofa performs similarly to Car on Sofa.

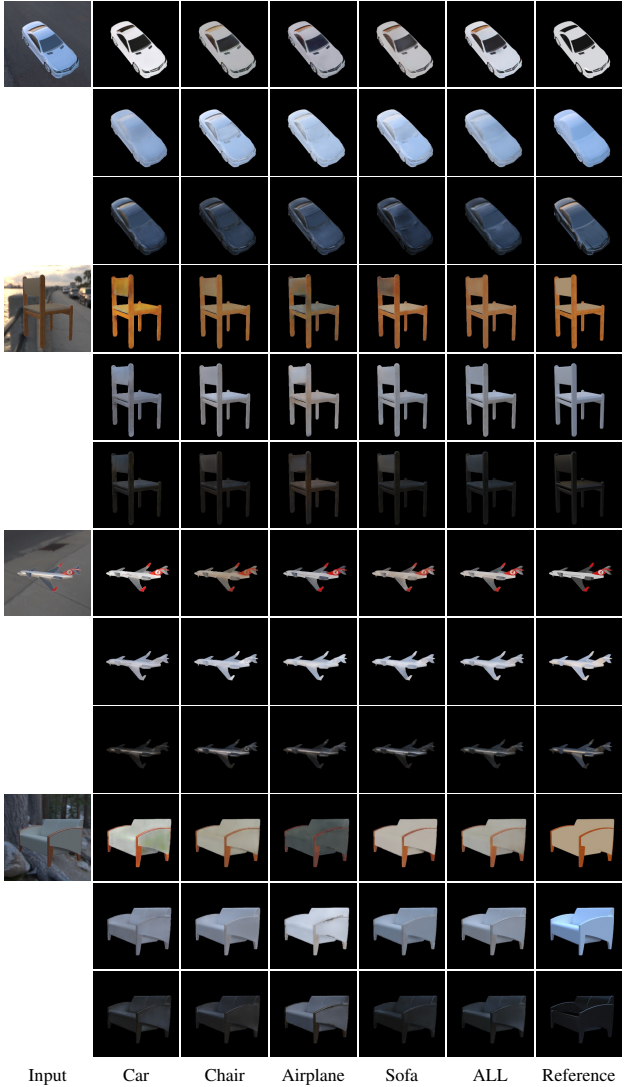

Input     Car      Chair    Airplane   Sofa     ALL      Reference

Figure 7: Cross-category comparison. Training on one specific category produces the best result when tested on objects from the same category. Categories with similar appearance also share similar results, *e.g.* sofas tested on the model trained on chairs. Dissimilar categories might produce results with artifacts, *e.g.* chairs tested on the model trained on airplanes.

network.

We test the role of our encoder and decoder in our image-to-image regression task, and verify which part is more critical for cross-category generalization. After training on a specific category, we freeze the encoder and fine tune the decoder on another category, e.g. we finetune the car-trained model on chairs, with the encoder fixed. The encoder features cannot be changed and we can only modify the way the decoder composes them. Table 4 shows the results on fine-tuned models. We observe that finetuning the decoder brings very limited improvement on the dataset it is fine tuned on, indicating that the encoder features are crucial to learning the decomposition. That fact that the model trained on ALL categories produces similar errors to category-specific models shows that the encoder of our model is able to capture both category-dependent and category-independent features.

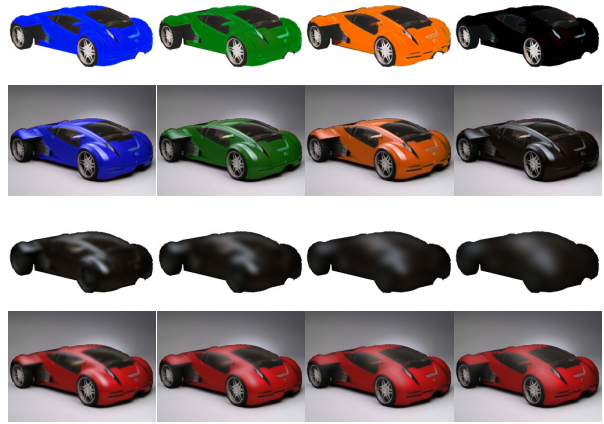

Figure 8: Image based appearance editing through intrinsic layers. The first and third row show the edited albedo and specular, by modifying results from Figure 6. The second and fourth row show the reconstructed images, delivering natural appearance in a scene.

## 7. Application

Figure 8 shows material editing examples based on our intrinsics results in Fig 6. We can recolor the diffuse albedo map to simulate a different color paint on the car, while preserving the shading and specular highlights. By blurring the specular component and changing its scale, we can also change the appearance of the car from highly specular to matte. These effects are hard to achieve on the original image but easily done on intrinsic layers without artifacts.

## 8. Conclusion

We reformulate the intrinsic image problem by introducing a specular term and solve this non-Lambertian intrinsics problem with a deep learning approach. A large scale training dataset with realistic images is generated using physically based rendering on the ShapeNet object repository. Our model consistently outperforms the state-of-the-art both visually and numerically, demonstrating cross-category generalization. Non-Lambertian intrinsics greatly extends Lambertian intrinsics to a much wider range of real images and real applications such as albedo and specular editing.

# References

[1] M. Aittala, T. Aila, and J. Lehtinen. Reflectance modeling by neural texture synthesis. *ACM Trans. Graph.*, 35(4):65:1–65:13, July 2016. 2

[2] J. Barron and J. Malik. Intrinsic scene properties from a single rgb-d image. In *Proceedings of the IEEE Conference on Computer Vision and Pattern Recognition*, pages 17–24, 2013. 2

[3] J. T. Barron and J. Malik. Shape, illumination, and reflectance from shading. *IEEE transactions on pattern analysis and machine intelligence*, 37(8):1670–1687, 2015. 1, 2, 4

[4] H. G. Barrow and J. M. Tenenbaum. Recovering intrinsic scene characteristics from images. *Computer Vision Systems*, pages 3–26, 1978. 2

[5] S. Bell, K. Bala, and N. Snavely. Intrinsic images in the wild. *ACM Transactions on Graphics (TOG)*, 33(4):159, 2014. 2, 4

[6] S. Bell, P. Upchurch, N. Snavely, and K. Bala. Material recognition in the wild with the materials in context database. In *IEEE Conference on Computer Vision and Pattern Recognition*, 2015. 2

[7] A. Bousseau, S. Paris, and F. Durand. User-assisted intrinsic images. In *ACM Transactions on Graphics (TOG)*, volume 28, page 130. ACM, 2009. 2

[8] D. J. Butler, J. Wulff, G. B. Stanley, and M. J. Black. A naturalistic open source movie for optical flow evaluation. In A. Fitzgibbon et al. (Eds.), editor, *European Conf. on Computer Vision (ECCV)*, Part IV, LNCS 7577, pages 611–625. Springer-Verlag, Oct. 2012. 2

[9] A. X. Chang, T. Funkhouser, L. Guibas, P. Hanrahan, Q. Huang, Z. Li, S. Savarese, M. Savva, S. Song, H. Su, J. Xiao, L. Yi, and F. Yu. ShapeNet: An Information-Rich 3D Model Repository. Technical Report arXiv:1512.03012 [cs.GR], Stanford University — Princeton University — Toyota Technological Institute at Chicago, 2015. 3, 4

[10] Q. Chen and V. Koltun. A simple model for intrinsic image decomposition with depth cues. In *Proceedings of the IEEE International Conference on Computer Vision*, pages 241–248, 2013. 2

[11] D. Eigen and R. Fergus. Predicting depth, surface normals and semantic labels with a common multi-scale convolutional architecture. In *Proceedings of the IEEE International Conference on Computer Vision*, pages 2650–2658, 2015. 2

[12] R. Grosse, M. K. Johnson, E. H. Adelson, and W. T. Freeman. Ground truth dataset and baseline evaluations for intrinsic image algorithms. In *Computer Vision, 2009 IEEE 12th International Conference on*, pages 2335–2342. IEEE, 2009. 2, 5

[13] S. Gupta, P. Arbeláez, R. Girshick, and J. Malik. Inferring 3d object pose in rgb-d images. *arXiv preprint arXiv:1502.04652*, 2015. 3

[14] K. He, X. Zhang, S. Ren, and J. Sun. Deep residual learning for image recognition. *CoRR*, abs/1512.03385, 2015. 2

[15] W. Jakob. Mitsuba renderer, 2010. http://www.mitsuba-renderer.org. 4

[16] J. T. Kajiya. The rendering equation. In *ACM Siggraph Computer Graphics*, volume 20, pages 143–150. ACM, 1986. 3

[17] E. P. Lafortune and Y. D. Willems. *Using the modified phong reflectance model for physically based rendering*. Citeseer, 1994. 4

[18] E. H. Land and J. J. McCann. Lightness and retinex theory. *JOSA*, 61(1):1–11, 1971. 2

[19] J. Liebelt and C. Schmid. Multi-view object class detection with a 3d geometric model. In *Computer Vision and Pattern Recognition (CVPR), 2010 IEEE Conference on*, pages 1688–1695. IEEE, 2010. 3

[20] T. Narihira, M. Maire, and S. X. Yu. Direct intrinsics: Learning albedo-shading decomposition by convolutional regression. In *Proceedings of the IEEE International Conference on Computer Vision*, pages 2992–2992, 2015. 1, 2, 4, 5, 6

[21] T. Narihira, M. Maire, and S. X. Yu. Learning lightness from human judgement on relative reflectance. In *IEEE Conference on Computer Vision and Pattern Recognition*, Boston, MA, 8-10 June 2015. 2

[22] D. Pathak, P. Kraehenbuehl, S. X. Yu, and T. Darrell. Constrained structured regression with convolutional neural networks. In *http://arxiv.org/abs/1511.07497*, 2016. 2

[23] X. Peng, B. Sun, K. Ali, and K. Saenko. Exploring invariances in deep convolutional neural networks using synthetic images. *arXiv preprint arXiv:1412.7122*, 2014. 3

[24] B. T. Phong. Illumination for computer generated pictures. *Communications of the ACM*, 18(6):311–317, 1975. 4

[25] K. Rematas, T. Ritschel, M. Fritz, E. Gavves, and T. Tuytelaars. Deep reflectance maps. In *The IEEE Conference on Computer Vision and Pattern Recognition (CVPR)*, June 2016. 2, 3, 4

[26] F. Romeiro and T. Zickler. Blind reflectometry. In *Proceedings of the 11th European conference on Computer vision: Part I*, ECCV'10, pages 45–58, Berlin, Heidelberg, 2010. Springer-Verlag. 3

[27] O. Ronneberger, P. Fischer, and T. Brox. *U-Net: Convolutional Networks for Biomedical Image Segmentation*, pages 234–241. Springer International Publishing, Cham, 2015. 2, 4

[28] C. Rother, M. Kiefel, L. Zhang, B. Schölkopf, and P. V. Gehler. Recovering intrinsic images with a global sparsity prior on reflectance. In *Advances in neural information processing systems*, pages 765–773, 2011. 2

[29] L. Shen, P. Tan, and S. Lin. Intrinsic image decomposition with non-local texture cues. In *Computer Vision and Pattern Recognition, 2008. CVPR 2008. IEEE Conference on*, pages 1–7. IEEE, 2008. 2

[30] L. Shen and C. Yeo. Intrinsic images decomposition using a local and global sparse representation of reflectance. In *Computer Vision and Pattern Recognition (CVPR), 2011 IEEE Conference on*, pages 697–704. IEEE, 2011. 2

[31] M. Stark, M. Goesele, and B. Schiele. Back to the future: Learning shape models from 3d cad data. In *BMVC*, volume 2, page 5, 2010. 3

[32] H. Su, C. R. Qi, Y. Li, and L. J. Guibas. Render for cnn: Viewpoint estimation in images using cnns trained with rendered 3d model views. In *Proceedings of the IEEE Inter-*

*national Conference on Computer Vision*, pages 2686–2694, 2015. 3

[33] Y. Weiss. Deriving intrinsic images from image sequences. In *Computer Vision, 2001. ICCV 2001. Proceedings. Eighth IEEE International Conference on*, volume 2, pages 68–75. IEEE, 2001. 2

[34] T. Zhou, P. Krähenbühl, and A. A. Efros. Learning data-driven reflectance priors for intrinsic image decomposition. *International Conference on Computer Vision*, abs/1510.02413, 2015. 2

[35] D. Zoran, P. Isola, D. Krishnan, and W. T. Freeman. Learning ordinal relationships for mid-level vision. In *International Conference on Computer Vision*, 2015. 2

# Learning Non-Lambertian Object Intrinsics across ShapeNet Categories
# Supplementary Material

Jian Shi
Institute of Software, Chinese Academy of Sciences
University of Chinese Academy of Sciences
shij@ios.ac.cn

Yue Dong
Microsoft Research Asia
yuedong@microsoft.com

Hao Su
Stanford University
haosu@cs.stanford.edu

Stella X. Yu
UC Berkeley / ICSI
stellayu@berkeley.edu

## 1. Network

We propose an encoder-decoder CNN for intrinsic image decomposition. The basic network structure is shown in Figure 1. Figure 1 shows the feature size in each encoder/decoder layer. The network design is mirrored with symmetric feature sizes for the encoder and the decoder, as well as 3-channel RGB input and output intrinsic components ($256 \times 256 \times 3$). To keep visual details and produce sharp outputs (*e.g.* CNNs as used by Direct Intrinsics [5] usually produce blurry results), we **link** early features in encoder layers to corresponding decoder layers of the same size (red arrows in the figure). Therefore, we call our network **Mirror-Link** CNN.

We build our network with simple layers. For the encoder, we only use a convolutional layer with $3 \times 3$ kernel and stride 2 to extract features for each level (every conv layer is followed by BN and ReLU). For the decoder, we first up-sample the previous layer's feature maps (unnecessary for the first decoder layer), and then concatenate them with their encoder counterpart. After that, features are passed through a Conv($3 \times 3$)-BN-ReLU sequence.

We use shared encoder and separate decoders, since intrinsic components (albedo, shading and specular) are not independent of each other. To further strengthen the correlation between albedo, shading and specular, we link the features across decoders (blue arrows in the figure).

Some other networks use the same idea of skip links, *e.g.* U-Net [7] and Deep Reflectance Map [6]. However, our network is different from theirs in following three aspects: We target a different and more complex image-to-image regression problem; our network is strictly symmetric in feature maps, including input and output; we have three outputs/decoders with a shared input/encoder.

We evaluate many variations of the network, including: 1) independent Mirror-Link network for albedo, shading and specular, 2) shared encoder without cross links (blue arrows) between decoders; 3) different numbers of skip links (red arrows) between corresponding encoder and decoder. Table 1 lists benchmark errors.

**Analysis.** By comparing *Independent* and *Shared Encoder* networks, we observe that although the latter contains fewer parameters, the performance is comparable for albedo and shading, even better for specular. Therefore, we consider a single shared encoder sufficient for extracting features for all 3 outputs.

As a multi-task CNN, there are some correlations among our 3 intrinsic components. Therefore, in *Mirror-Link* network, we further strengthen such correlations by cross links in the decoders. The performance is improved significantly.

To evaluate the skip links from the encoder to the decoder, we test *Skip Link-3* and *Skip Link-0*. In *Skip Link-*

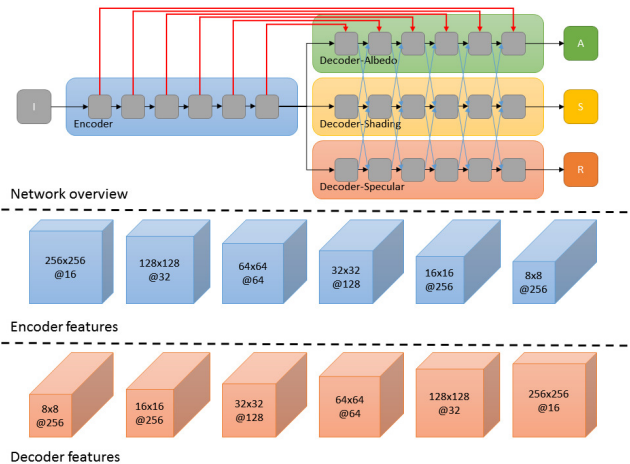

Figure 1: Network Structure.

|  | Albedo | | | Shading | | | Specular | | |
| --- | --- | --- | --- | --- | --- | --- | --- | --- | --- |
|  | MSE | LMSE | DSSIM | MSE | LMSE | DSSIM | MSE | LMSE | DSSIM |
| Independent | 0.0116 | 0.0469 | 0.1183 | 0.0059 | 0.0111 | 0.0667 | 0.0064 | 0.0845 | 0.1179 |
| Shared Encoder | 0.0122 | 0.0503 | 0.1210 | 0.0069 | 0.0124 | 0.0746 | 0.0058 | 0.0767 | 0.1077 |
| **Mirror-Link** | **0.0083** | **0.0353** | **0.0939** | **0.0055** | **0.0097** | **0.0622** | **0.0042** | **0.0578** | **0.0831** |
| Skip Links - 3 | 0.0127 | 0.0527 | 0.1282 | 0.0079 | 0.0149 | 0.0815 | 0.0059 | 0.0858 | 0.1152 |
| Skip Links - 0 | 0.0226 | 0.0794 | 0.1705 | 0.0118 | 0.0225 | 0.0977 | 0.0084 | 0.1234 | 0.1448 |

Table 1: Numeric comparison for variant network structures. *Independent* uses 3 independent encoder-decoder networks for albedo, shading and specular, nothing is shared. *Shared Encoder* uses the shared encoder and separate decoders, but without cross links(blue arrows in the figure) between decoders. *Skip Link-0* is the network without skip links (red arrows) between encoder and decoder. *Skip Link-3* uses 3 links in the middle. *Mirror-Link* contains all links and is the model we used in this paper. We use the same metrics as [5], where MSE is a scale-invariant version.

*3*, we remove 3 links outside while keep 3 in the middle. In *Skip Link-0*, we remove all the skip links (red arrows in the figure). We observe that skip links improve the performance.

## 2. Loss Functions

Similar to [5], a scale-invariant MSE(SMSE) loss, combined with standard MSE loss, is employed in our work. The SMSE first scales the predicted output and then compares MSE with the groundtruth.

$$SMSE(X, X_{gt}) = MSE(\alpha X, X_{gt}) \quad (1)$$

$$\alpha = \text{argmin} \, MSE(\alpha X, X_{gt}) \quad (2)$$

Previous works assume that $I = A \times S$, making the scaling ambiguity in $\alpha A$ and $\frac{1}{\alpha} S$. In our formulation for non-Lambertian objects as $I = A \times S + R$, we also have the same scaling ambiguity for $A \times S$. Thus, we apply the scale-invariant loss for albedo and shading. However, for the specular $R$, either scale-invariant ($\alpha R$) or shift-invariant ($\alpha + R$) would bring different patterns to $A \times S$. Thus, we simply apply MSE loss for the specular output. Since we only have ground-truth for objects, we use masks for background pixels for computing the loss and back-propagating gradients.

For albedo and shading, we use

$$E_{A,S} = 0.95 \times SMSE + 0.05 \times MSE, \quad (3)$$

and for specular, we use:

$$E_R = MSE. \quad (4)$$

## 3. Rendering Pipeline

There are about $57,000$ models in the ShapeNet core database [3]. Most of them have materials and textures. All these models are normalized and aligned to a unit bounding box. Due to computing resource limitations, we only render $31,072$ models, more than a half of the database.

A physics-based open source render Mitsuba [4] is employed for the rendering task. It can directly render the input image and the groundtruth albedo. For the shading component, we replace materials with pure diffuse white for rendering. For the specular, we set the diffuse to $0$ and keep the specular for rendering.

Models are rendered under all 98 environment maps we have. Viewpoints are randomly assigned on the upper hemisphere for the object in each environment map. We use a low discrepancy Halton sequence to generate random viewpoints to keep them uniform yet random in the distribution.

We use path tracing to render images. To reduce the rendering time, we render albedo, shading and specular (as well as an object mask) and synthesize the image by $I = A \times S + R$. It saves us $30\%$ of rendering time. Albedo and mask rendering is simple and fast, while image, shading and specular require many more samples for path tracing.

## 4. More Results

Pages 3-6 show some results from our synthetic testing set. Groundtruth (reference) is included for visual comparisons. Figure 2 shows results on real images downloaded from the Internet. Results from Direct Intrinsics [5] (both their released model and the model trained on our dataset) and SIRFS [1] are provided for visual comparisons.

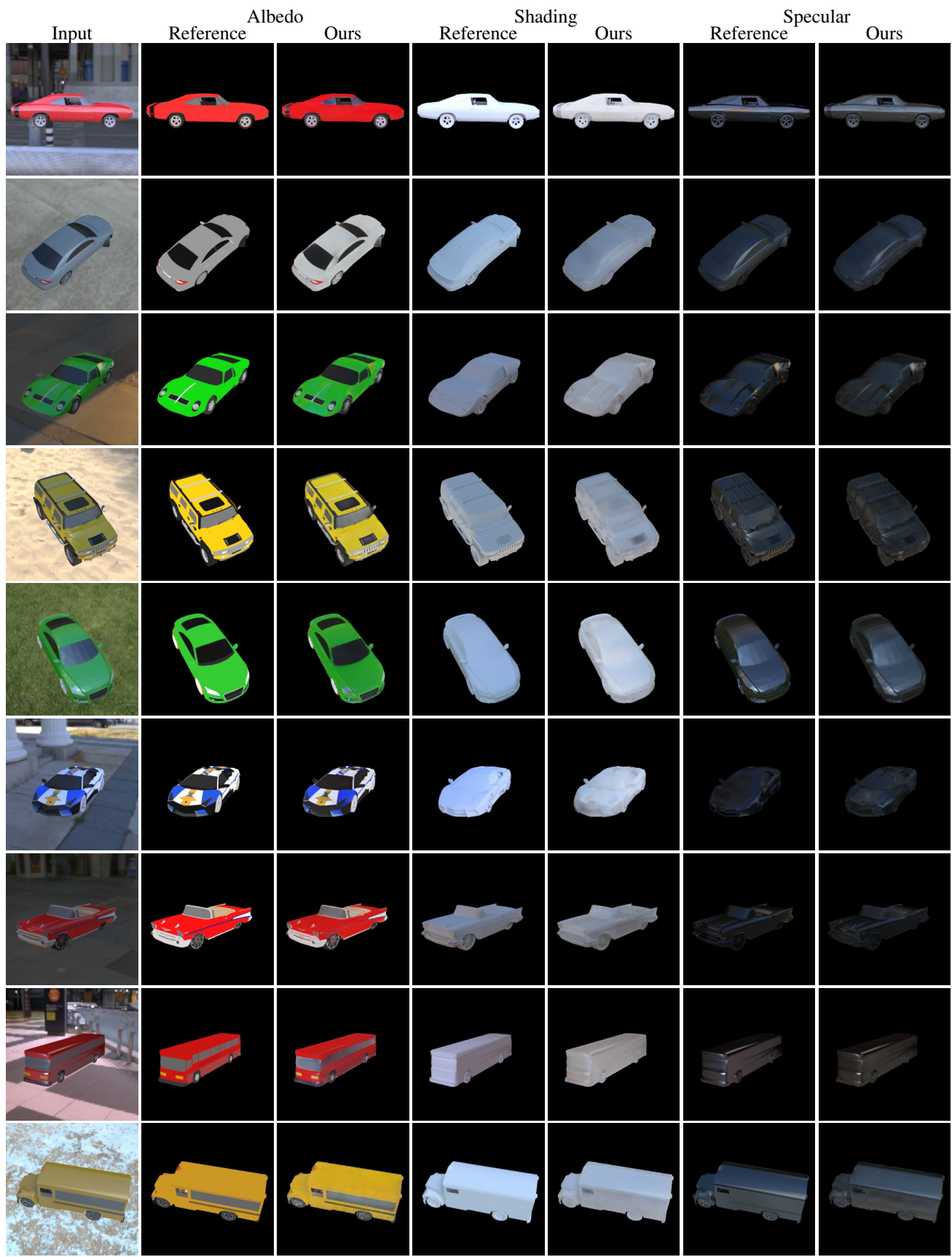

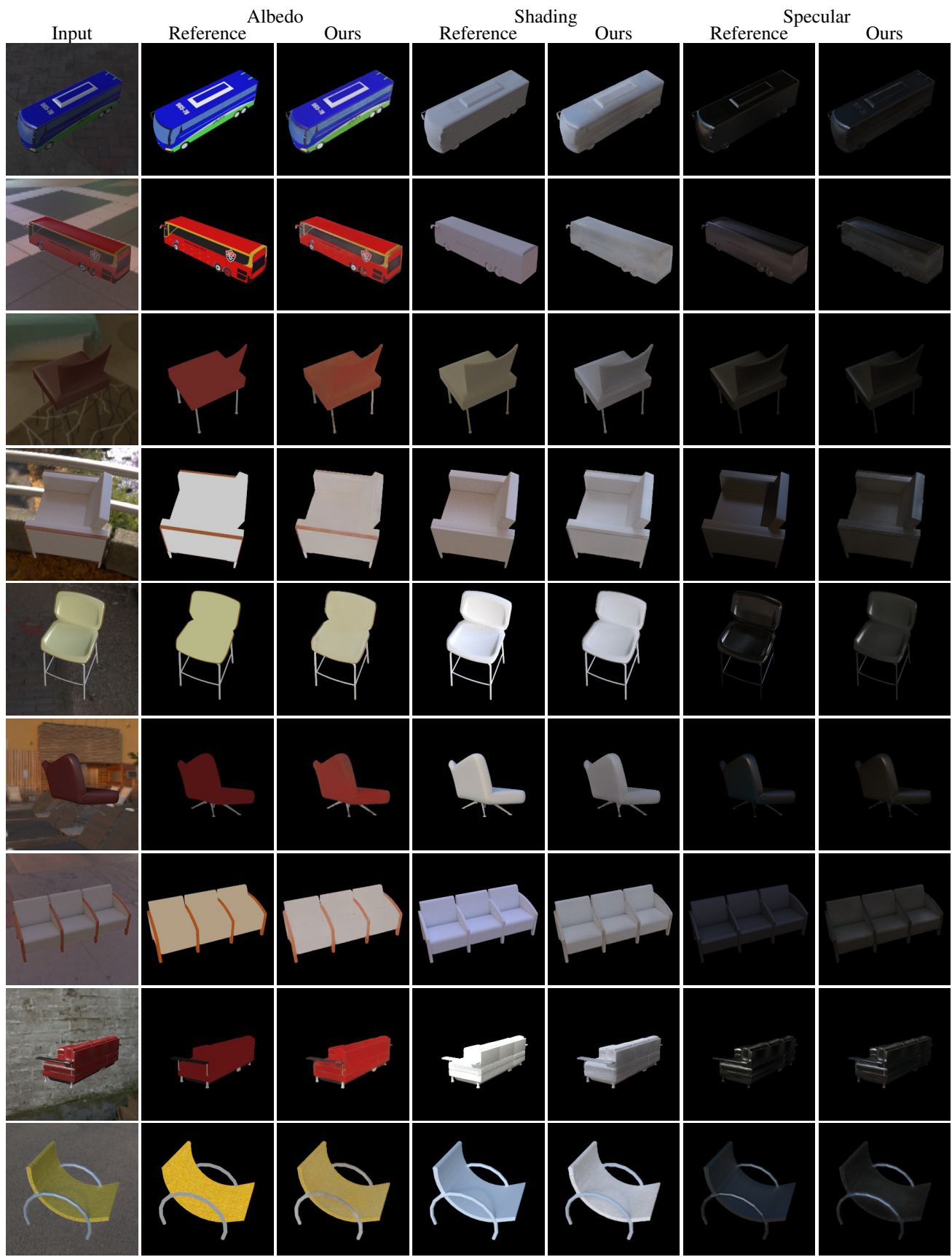

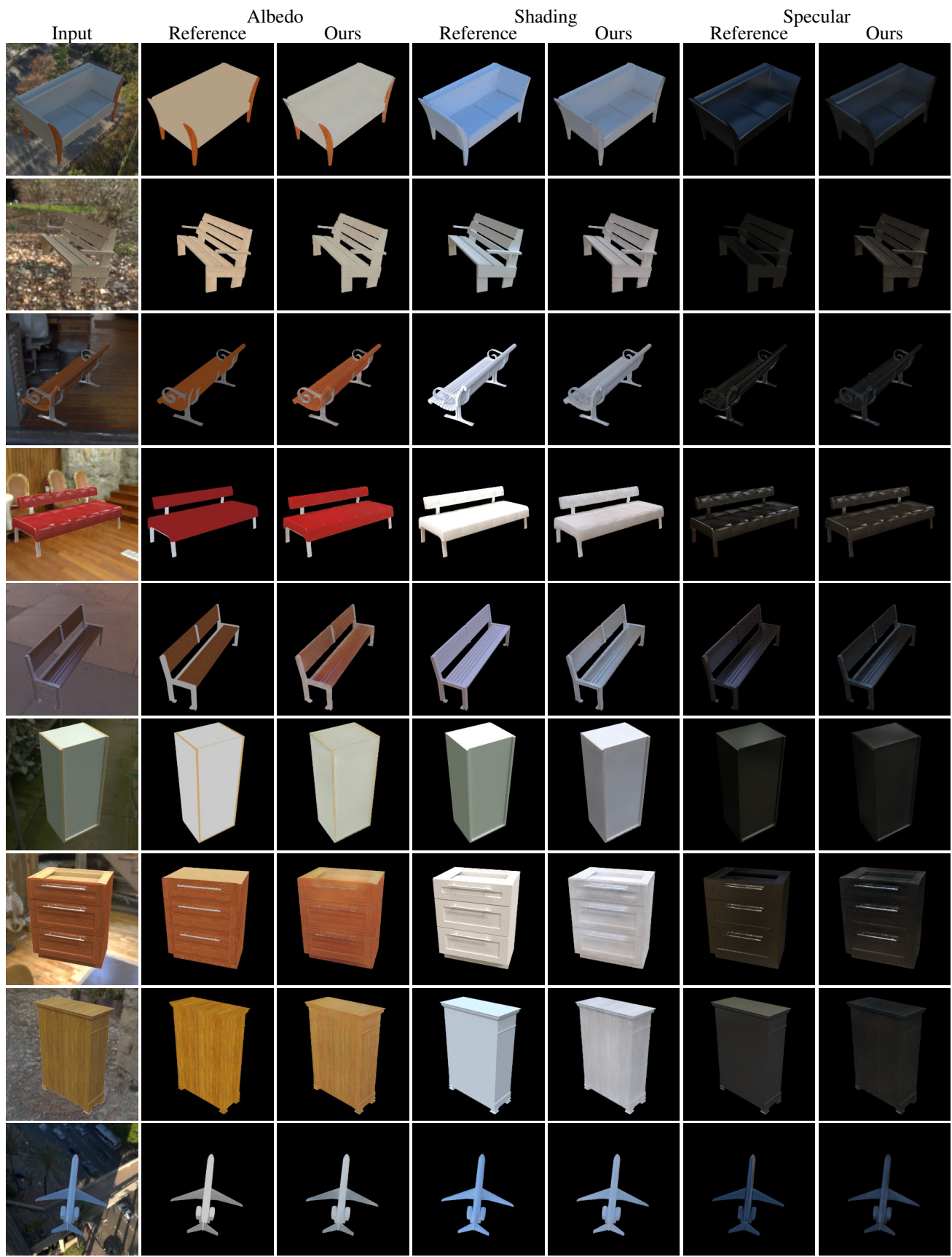

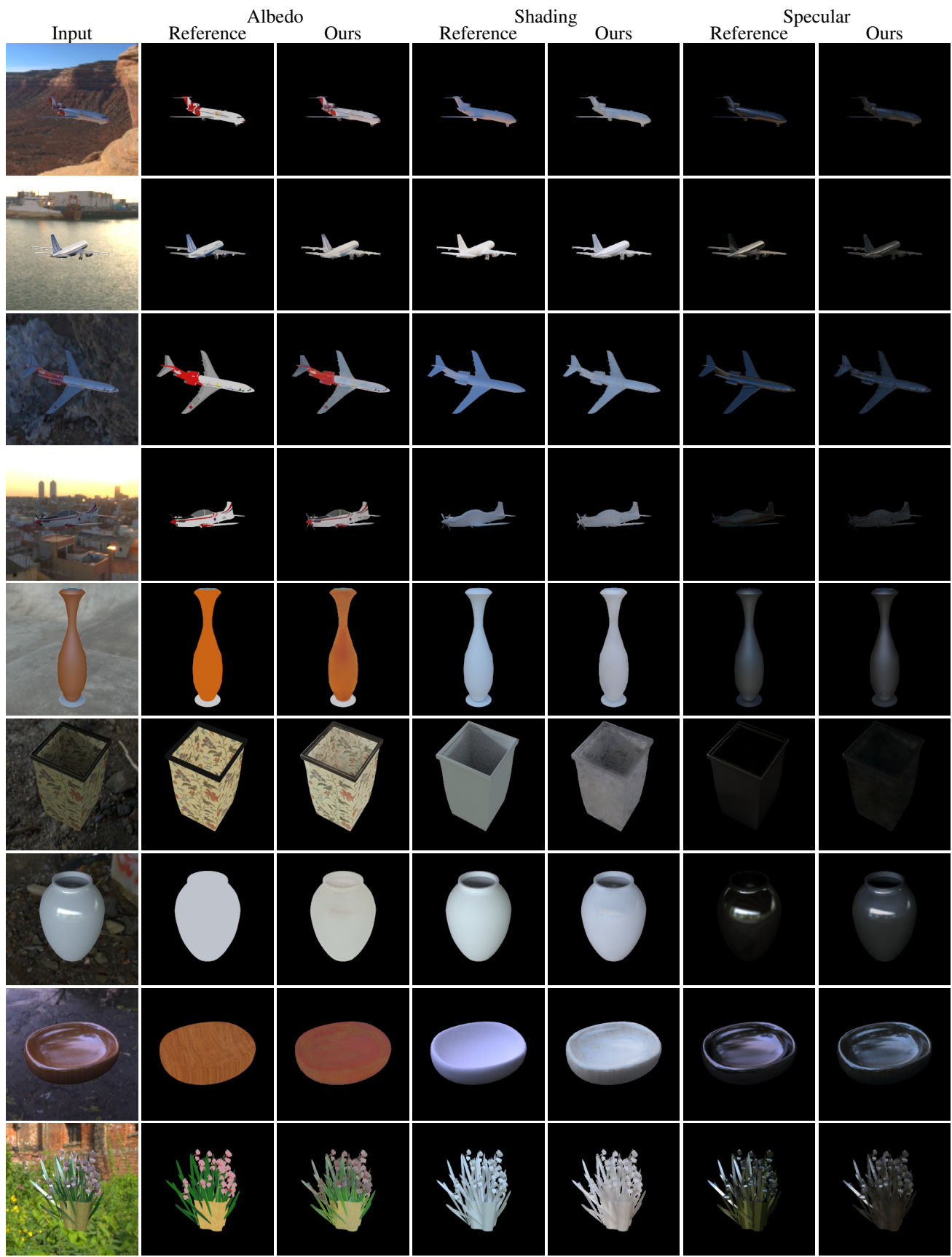

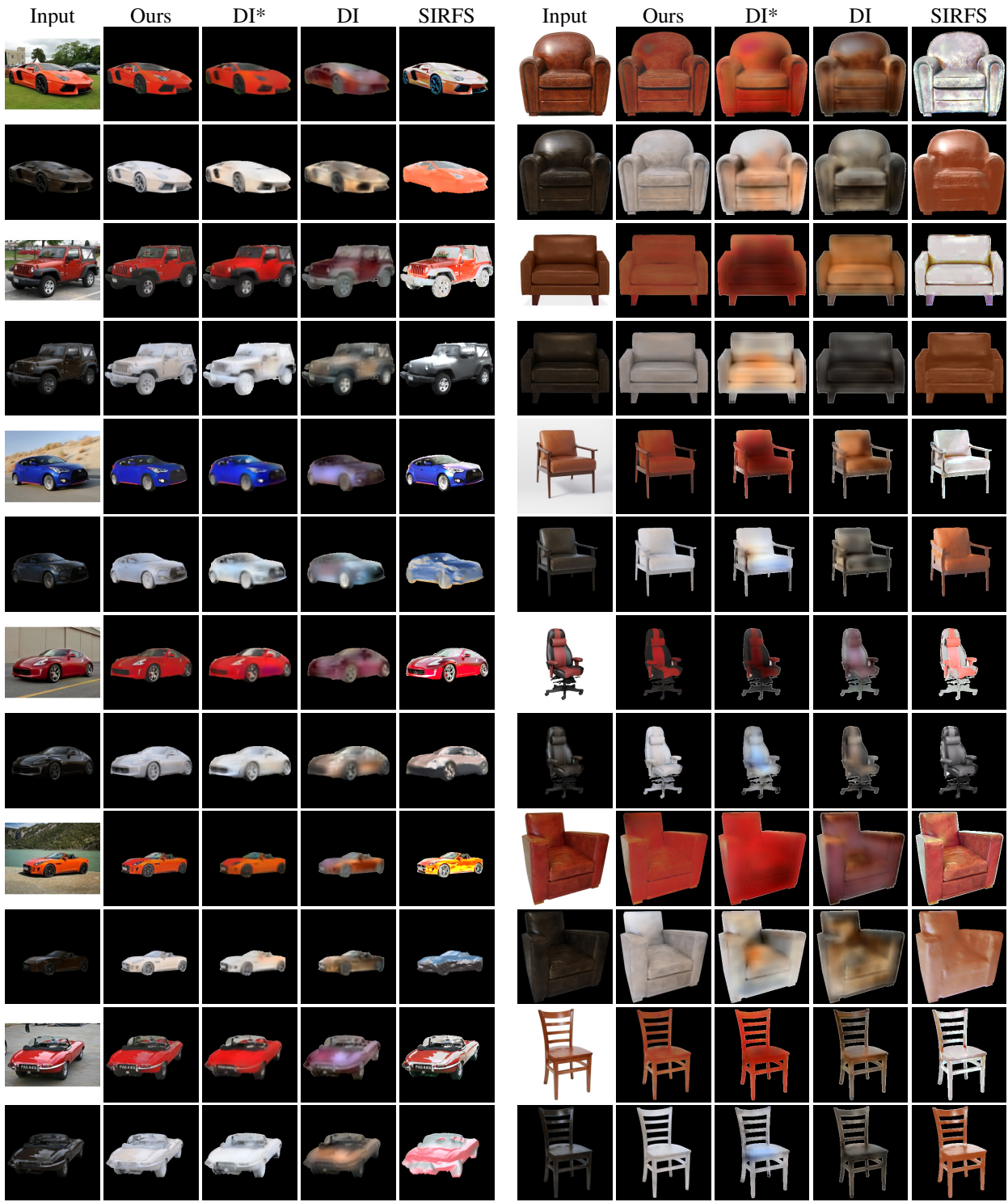

Figure 2: Real image results from our method, SIRFS, Direct Intrinsics(DI) and Direct Intrinsics model trained on our dataset(DI*). The first column shows input image(top) and specular solved by our model(bottom). Other columns are albedo(top) and shading(bottom) results. We will publish our trained model as well as the synthetic dataset. For more results, please try our released model.

# 5. Another synthetic evaluation

We build an object-level intrinsic image dataset based on ShapeNet. Objects are rendered within real captured environment maps to approximate realistic illuminations. For more realistic and advanced effects such as occlusion and inter-reflection, rendering objects within 3D scenes is preferred. Figure 3 shows a photo-realistic rendered scene [2]. Unfortunately, there is no existing large-scale public realistic scene database for generating training data. However, we can still make use of those limited number of public 3D scenes to generate groundtruth data for evaluation.

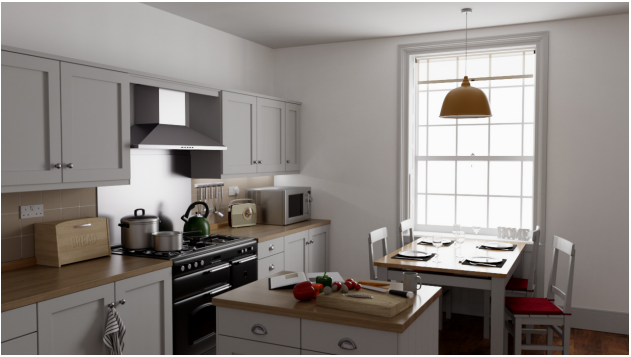

Figure 3: A rendered kitchen scene.

## 5.1. Pepper

The red and orange peppers on the table are rendered. There are sharp specular highlights, as well as inter-reflections from nearby objects. Figure 4 shows sample results. Table 2 lists estimation errors. Although there is no similar object in our training set, our method produces acceptable results especially for shading and specular.

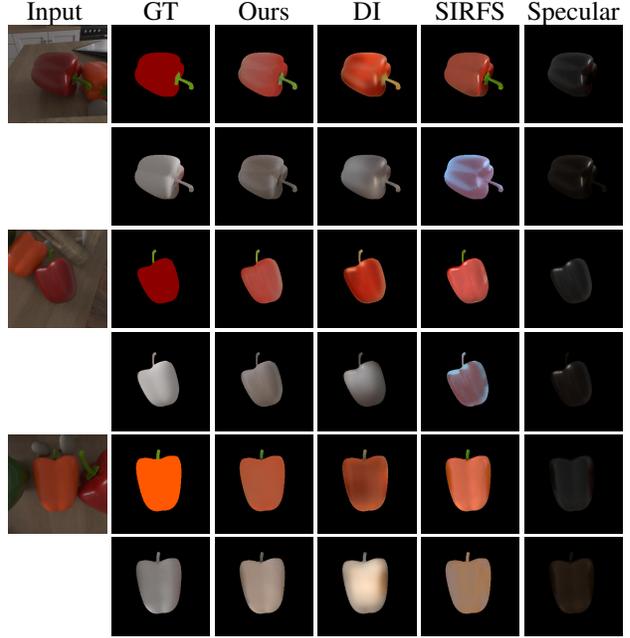

Figure 4: Visual results for the red pepper. The last column shows the groundtruth specular(top) and ours(bottom).

## 5.2. Kettle

The green metal kettle is also rendered for evaluation. In this case, the specular is not as sharp as the pepper, but with larger area. Figure 5 shows the visual results. Our algorithm produces reasonable specular and shading. For the albedo component, although our model failed to fill the correct color for the holes on the specular area, it still produces better results than others. Table 3 shows the numeric errors.

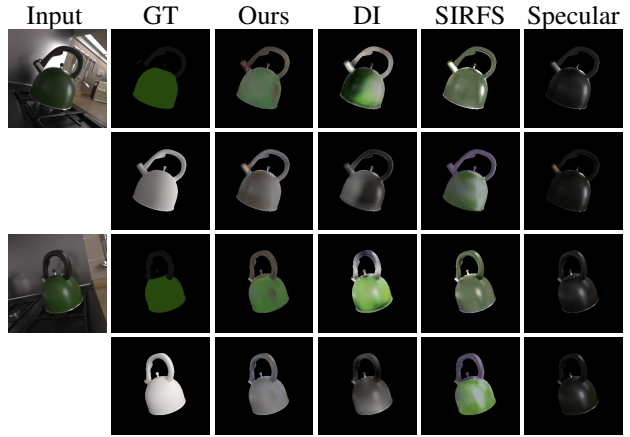

Figure 5: Visual results for the green kettle. The last column shows the groundtruth specular(top) and ours(bottom).

|  |  | MSE | | LMSE | | DSSIM | |
|---|---|---|---|---|---|---|---|
|  |  | albedo | shading | albedo | shading | albedo | shading |
| Pepper-1 | SIRFS | 0.0078 | 0.0110 | 0.1294 | 0.0179 | 0.3287 | 0.0959 |
| | DI | 0.0083 | 0.0093 | 0.0909 | 0.0091 | 0.3345 | 0.0530 |
| | Ours | **0.0058** | **0.0030** | **0.0883** | **0.0040** | **0.3219** | **0.0446** |
| | specular | 0.0012 | | 0.0162 | | 0.0776 | |
| Pepper-2 | SIRFS | 0.0032 | 0.0255 | 0.1283 | 0.0146 | 0.3491 | 0.1054 |
| | DI | 0.0033 | 0.0147 | 0.0940 | 0.0052 | **0.3293** | 0.0760 |
| | Ours | **0.0022** | **0.0023** | **0.0789** | **0.0017** | 0.3302 | **0.0544** |
| | specular | 0.0009 | | 0.0114 | | 0.0636 | |
| Pepper-3 | SIRFS | 0.0037 | 0.0033 | 0.1322 | 0.0035 | 0.1901 | 0.0308 |
| | DI | 0.0064 | 0.0077 | **0.0743** | 0.0055 | 0.2037 | 0.0682 |
| | Ours | **0.0006** | **0.0009** | 0.0963 | **0.0016** | **0.1789** | **0.0165** |
| | specular | 0.0002 | | 0.0090 | | 0.0300 | |

Table 2: Error comparison for peppers.

|  |  | MSE | | LMSE | | DSSIM | |
| --- | --- | --- | --- | --- | --- | --- | --- |
|  |  | albedo | shading | albedo | shading | albedo | shading |
| Kettle-1 | SIRFS | 0.0076 | 0.0128 | 0.0890 | 0.0171 | 0.2912 | 0.0943 |
|  | DI | 0.0112 | 0.0679 | 0.1275 | 0.0576 | 0.2882 | 0.1945 |
|  | Ours | **0.0027** | **0.0063** | **0.0310** | **0.0103** | **0.2297** | **0.0737** |
|  | specular | 0.0015 | | 0.0255 | | 0.0439 | |
| Kettle-2 | SIRFS | 0.0060 | 0.0307 | 0.0777 | 0.0190 | 0.2851 | 0.1592 |
|  | DI | 0.0042 | 0.0717 | 0.0508 | 0.0354 | 0.3295 | 0.2329 |
|  | Ours | **0.0027** | **0.0101** | **0.0326** | **0.0091** | **0.2049** | **0.1254** |
|  | specular | 0.0009 | | 0.0100 | | 0.0273 | |

Table 3: Error comparison for the kettle.

## 6. Intrinsic Video

We apply our model to each frame in a video, without any temporal consistency constraints across frames. It produces stable and reasonable results.

## References

[1] J. T. Barron and J. Malik. Shape, illumination, and reflectance from shading. *IEEE transactions on pattern analysis and machine intelligence*, 37(8):1670–1687, 2015. 2

[2] B. Bitterli. Rendering resources, 2016. https://benedikt-bitterli.me/resources/. 8

[3] A. X. Chang, T. Funkhouser, L. Guibas, P. Hanrahan, Q. Huang, Z. Li, S. Savarese, M. Savva, S. Song, H. Su, J. Xiao, L. Yi, and F. Yu. ShapeNet: An Information-Rich 3D Model Repository. Technical Report arXiv:1512.03012 [cs.GR], Stanford University — Princeton University — Toyota Technological Institute at Chicago, 2015. 2

[4] W. Jakob. Mitsuba renderer, 2010. http://www.mitsuba-renderer.org. 2

[5] T. Narihira, M. Maire, and S. X. Yu. Direct intrinsics: Learning albedo-shading decomposition by convolutional regression. In *Proceedings of the IEEE International Conference on Computer Vision*, pages 2992–2992, 2015. 1, 2

[6] K. Rematas, T. Ritschel, M. Fritz, E. Gavves, and T. Tuytelaars. Deep reflectance maps. In *The IEEE Conference on Computer Vision and Pattern Recognition (CVPR)*, June 2016. 1

[7] O. Ronneberger, P. Fischer, and T. Brox. *U-Net: Convolutional Networks for Biomedical Image Segmentation*, pages 234–241. Springer International Publishing, Cham, 2015. 1